\documentclass[conference]{IEEEtran}
\usepackage{pifont}  
\usepackage{array}   
\IEEEoverridecommandlockouts
\usepackage{cite}
\usepackage{amsmath,amssymb,amsfonts}
\usepackage{algorithm}
\usepackage{algcompatible}
\usepackage{graphicx}
\usepackage{textcomp}
\usepackage{float}
\usepackage{xcolor}
\def\BibTeX{{\rm B\kern-.05em{\sc i\kern-.025em b}\kern-.08em
    T\kern-.1667em\lower.7ex\hbox{E}\kern-.125emX}}
\begin{document}

\title{Autonomous Crack Detection using Deep Learning on Synthetic Thermogram Datasets
%
}
\author{
\IEEEauthorblockN{Chinmay Makarand Pimpalkhare\IEEEauthorrefmark{1}\textsuperscript{1} and D. N. Pawaskar,\textsuperscript{1}}
\IEEEauthorblockA{\textsuperscript{1}\textit{Department of Mechanical Engineering,} \\
\textit{Indian Institute of Technology Bombay, Mumbai - 400076, India}}
}

\maketitle
\footnote{\IEEEauthorrefmark{1}
Corresponding Author: chinmaymrpimpalkhare@gmail.com}
\begin{abstract}
In a lot of scientific problems, there is the need to generate data through the running of an extensive number of experiments. Further, some tasks require constant human intervention. We consider the problem of crack detection in steel plates. The way in which this generally happens is how humans looking at an image of the thermogram generated by heating the plate and classifying whether it is cracked or not. There has been a rise in the use of Artificial Intelligence (AI) based methods which try to remove the requirement of a human from this loop by using algorithms such as Convolutional Neural Netowrks (CNN)s as a proxy for the detection process. The issue generally is that CNNs and other vision models are generally very data-hungry and require huge amounts of data before they can start performing well. This data generation process is not very easy and requires innovation in terms of mechanical and electronic design of the experimental setup. It further requires massive amount of time and energy, which is difficult in resource-constrained scenarios. We try to solve exactly this problem, by creating a synthetic data generation pipeline based on Finite Element Simulations. We employ data augmentation techniques on this data to further increase the volume and diversity of data generated. The working of this concept is shown via performing inference on fine-tuned vision models and we have also validated the results by checking if our approach translates to realistic experimental data. We show the conditions where this translation is successful and how we can go about achieving that. 
\end{abstract}


\section{Introduction}
\subsection{Need for Automated Crack Detection}

Failure in materials can occur due to the propagation of local cracks, which have the potential to cause widespread issues throughout the material. This susceptibility to crack propagation is often exacerbated by factors such as manufacturing errors or defects. A notable consequence of crack propagation is the increased vulnerability of the material to corrosion.
\par 
The problem of reducing the possibilities of cracks naturally leads to the question: "How can we effectively detect these cracks as soon as possible?". A lot of times, sufficient manpower is not available to perform the tasks of checking for cracks. In such cases, automated systems may be employed. These automated systems generally employ computer vision algorithms in the background. They can effectively match the performance of human annotators. 

\subsection{Requirement of Large-scale Datasets} The deep learning automated systems require massive amounts of data to be trained. Further, this data needs to be diverse so that a system trained on this data can scale and adapt to new scenarios. This process in known as domain adaptation. The issue is that it is very difficult to generate such large-scale datasets. Most Non-Destructive Testing (NDT) datasets, which are generated using techniques such as thermography are not available publicly. There are a large number of agreements because a majority of such datasets are related to matters of defence and thus there is a lack of open-source datasets. Further, if one were to produce such a dataset via experimentation in a lab, it would be a task requiring massive amounts of time and energy. Thermal experiments generally need a lot of energy due to the high specific heat of the metals involved. In addition to that, to reach the high temperatures takes a lot of time. This leads us to an issue where we might have to train neural networks on scarce data, which leads to drastic decrease in their performance. There is henceforth a need of implementing synthetic data generation methods that can leverage the techniques offered by simulation models to create surrogate datasets that are very similar to the real datasets semantically. 
\section{Relevant Work}
\subsection{Synthetic Data Generation for Object Detection}
A lot of approaches in the past have applied synthetic data generation methods succesfully in various contexts. Jain et al. (2022) use Generative Adversarial Networks for generating data to train CNN models and use them for the task of surface crack detection. They report strong results through this process. They also show high generalization capacities for translation to manual inspection tasks. Wood et al. (2021) use synthetically generated face samples in their paper. They make use of no realistic images in their training procedure. He et al. (2022) consider the generative models to create their data. They try to answer the question of whether such an approach is good enough for image recognition tasks. Wang et al. (2021) consider images that have first been collected through a game and then reconstruct crowd depiction samples through this dataset. Hahner et al. (2021) create a simulation pipeline that can add weather effects to a variety of LIDAR based images. They focus on adding patterns such as foggy weather to any image. In their survey paper, Lu et al. (2023) consider various synthetic data generation to machine learning contexts. This will include both supervised and unsupervised settings. Roberts et al. (2021) create a pipeline for the understanding in indoor images through their pipeline called HyperSim which gives access to 3D data.  Tokmakev et al. (2021) also consider a dataset where they are considering the tracking of multiple objects. 
\subsection{Deep Learning for Crack Detection}

Various studies have addressed the critical task of crack detection using diverse methodologies and datasets. Yang, Wang et al. (2019) conducted experiments employing varying heat flux and thermal imagery, utilizing a Faster Region Proposal based Convolutional Neural Network for Object Detection. Similarly, Jaeger, Schmid et al. (2022) focused on turbine blade data, employing multiple deep learning models to classify infrared thermal images of turbine blades with cracks. In contrast, Mohan, Poobal (2018) provided a comprehensive review of crack detection methods, predominantly emphasizing vanilla image processing techniques without delving into deep learning.

Tian, Wang et al. (2021) explored a novel approach by implementing Generative Adversarial Networks (GANs) and Principal Component Analysis (PCA) for feature extraction in crack detection, emphasizing their feasibility for image augmentation with experimental data. Alexander, Hoskere et al. (2022) adopted the RFTNet, a Semantic Segmentation Network, specifically tailored for the fusion of RGB and thermal images in automated deep learning crack detection for civil architecture. Meanwhile, Chandra, AlMansoor et al. (2022) delved into the analysis of infrared thermal images of complex pavement defect conditions, incorporating seasonal effects using a vanilla Convolutional Neural Network.

Kovacs et al. (2020) explored deep learning approaches for thermographic imaging, employing synthetic data on virtual waves for Non-Destructive Testing. Fang et al. (2021) contributed to the field by using both synthetic and experimental data, employing a deep learning algorithm with pulsed thermography for automatic defects segmentation and identification, particularly focusing on simulations with CFRP materials. Wu et al. (2021) use a U-Net based architecture for the detection of small objects, which can include cracks in an image as well.

\section{Proposed Method}
\subsection{Overview}
\begin{figure}[H]
    \centering
\includegraphics[width 
 = 0.45\textwidth] {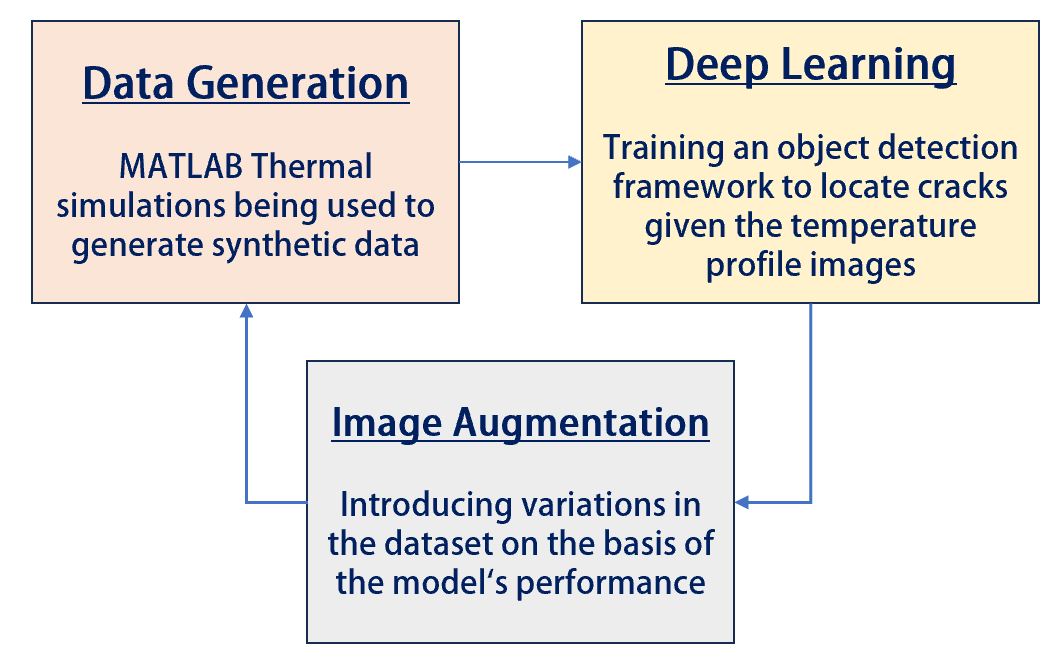}
    \caption{Overview of our solution}
    \label{fig:enter-label}
\end{figure}
Our approach involves three important aspects. The first is to create a data generation pipeline utilizing finite element simulations. The code has been written in MATLAB and has been automated such that it is possible to generate data quickly. At the moment, we are able to generate 100 images in 3 minutes, which is fast enough for most application since this data can be created and stored in some sort of buffer. The next crucial part is that of deep learning. The annotated images can be passed through the neural network and its performance can be analysed. The last part is that of image augmentation. This is an important part because it acts in a closed feedback loop with the data generation pipeline. The results of the deep learning framework make it clear "How the system is failing". Hence it is possible to generate examples which may act in an adversarial fashion. Modifying the simulation pipeline and enabling it to be able to detect cracks even in such hard scenarios is what finally makes the system robust to various settings. 

\subsection{Data Generation Pipeline}
\begin{figure}[H]
    \centering
\includegraphics[width 
 = 0.45\textwidth] {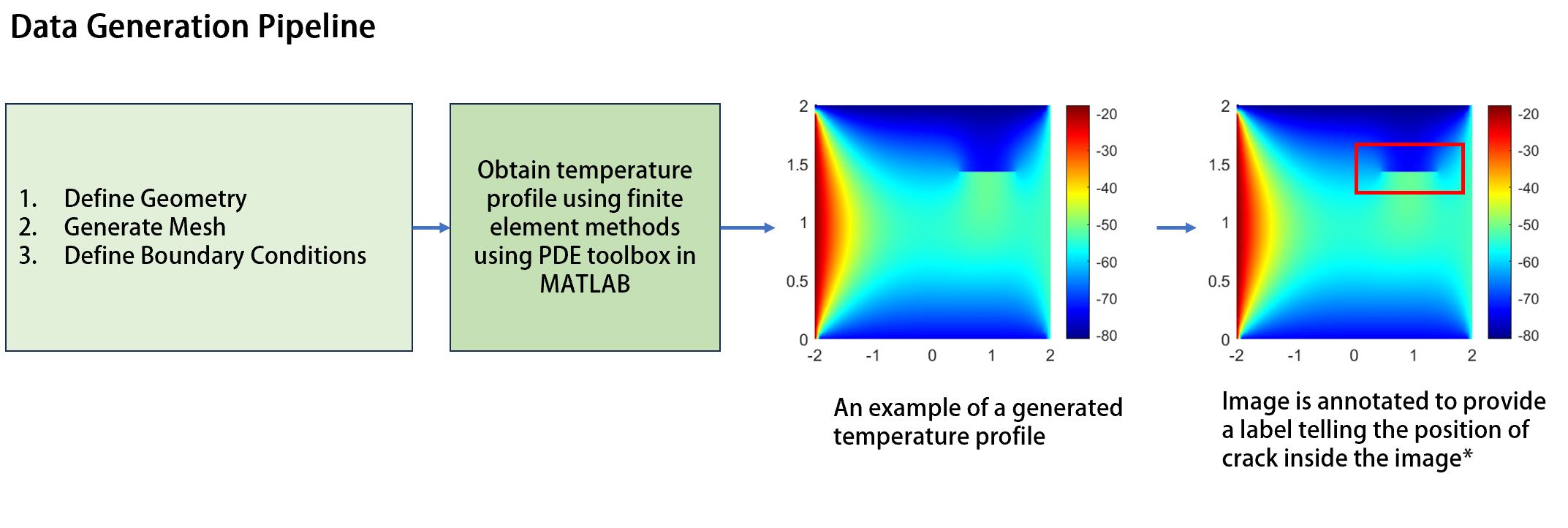}
    \caption{The Data Generation Pipeline}
    \label{fig:enter-label}
\end{figure}
In this section, we shall breifly explain the data generation pipeline and also discuss why specific design choices were made. The data which we are using is being generated through finite element simulations. 
The finite element method (FEM) is a popular method for numerically solving differential equations arising in engineering and mathematical modeling. It is a generally used for solving partial differential equations in two or three space variables. To solve a problem, the FEM subdivides a large system into smaller, simpler parts called finite elements. This is achieved by discretization in the dimensions, which is implemented by the construction of a mesh of the object, which is the numerical domain for the solution, which has a finite number of points. The finite element method formulation of a boundary value problem finally results in a system of algebraic equations. The method approximates the unknown function over the domain.[2] The simple equations that model these finite elements are then assembled into a larger system of equations that models the entire problem. The FEM then approximates a solution by minimizing an associated error function via the calculus of variations.
Annotations for crack detection are carried out using the CVAT.io annotation tool, enabling the precise marking of crack locations. The annotated information, specifying the location of cracks, is then stored in a text file, serving as input for the training of the algorithm. Additionally, boundary conditions provided during the annotation process are also stored.
\begin{figure}[H]
    \centering
\includegraphics[width 
 = 0.42\textwidth] {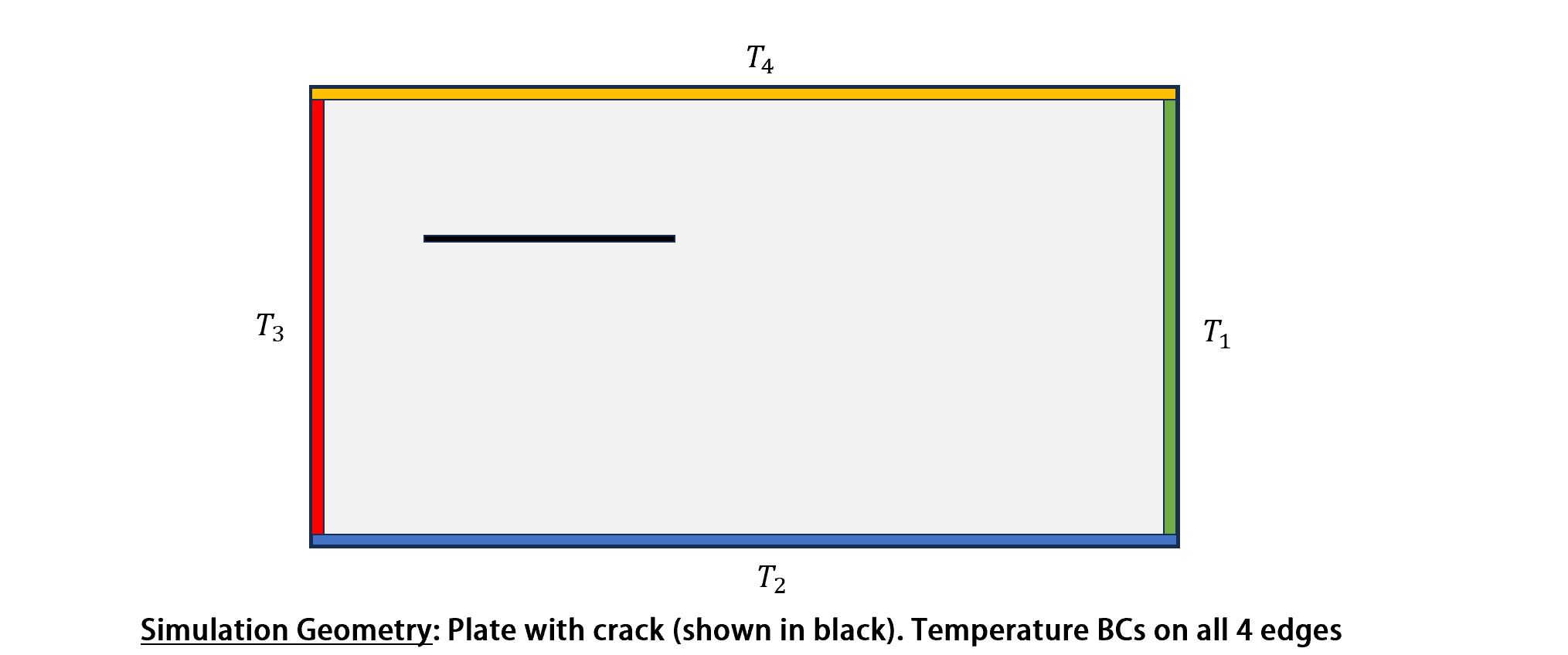}
    \caption{Simulation Template}
    \label{fig:enter-label}
\end{figure}
\par \noindent The first design choice is regarding the use of the base geometry. Since most of the objects we would study were in the form of plates, we decided to choose a rectangular outer boundary for the object. The outer boundary is not very important, and a circular one could be used too. However, using a polygonal boundary allows us to effectively define a single boundary condition at every side of the polygon, which is something that can be quickly randomized. 
\par \noindent The boundary conditions on the polygon's sides can be chosen from between a constant temperature boundary condition and a constant heat flux boundary condition. This was a base choice and it is very easy to implement more complicated and even time-varying boundary conditions within the framework which we have. The final focus lies on the temperature profile that is generated at steady state. The paper by Jaeger et al. shows that using specific kinds of boundary condition leads to more effective types of crack detection due to the thermal gradients aligning perpendicularly with the crack's direction, and this is what we successfully observe through our simulations too. 
\par \noindent The definition of the crack is very important since it has deep effects on the final outputs generated. We define a crack as a region where there is a loss in the base material. Hence, we remove the crack and for all practical purposes the crack is an empty space for us. This leads us to the question of choosing a correct and optimum value for the thickness of a crack. The thinner the crack, the less it will perturb the thermal profile around it and the harder it will be to detect. A thicker crack is easier to detect even for a human. However, it is good to use a mixture of both thin and thick cracks during the training procedure since the thin cracks act as a hard example while the thick cracks can allow the neural network to learn more about the local perturbations formed at the edges. 
\par \noindent In the later stages of our approach, also allow for the possibility of defining separate boundary conditions at the edge of the cracks in addition to the outer edges. These situations give rise to completely different scenarios which helps add to the diversity of the data. Just like the crack width, the crack length is also a parameter which we vary. 
\par \noindent An important part of creating the randomization is to create scenarios in which the geometry of the cracks change. This involves creating datasets where location of the crack and its orientation with the edges of the plates change. This is not just something which can be controlled through data augmentation methods such as cropping because we observe changes in the contour profiles generated when the orientation with the contour along which edge temperatures are constant is changed. 
\par \noindent We also consider scenario where there are multiple cracks in the same plate. Further, we allow for the possibility that both the cracks have completely different parameters. 
\par \noindent For the material properties, we have selected the material to be steel and defined the thermal conductivity, specific heat and other properties correspondingly. It would be a very interesting exercise to try out what happens in case of alloys or even materials which are fused together, because this will lead to situations where there are abrupt changes in the boundary conditions. 
\par \noindent The last part is choosing a colormap for the temperature profile so that it matches the result of a thermography image generated by a simulation. This was achieved by experimenting with different kinds of colormaps available in MATLAB. We found the 'jet' and 'inferno' colormaps to be particularly interesting and realistic. We have also used grayscale images by selecting the colormap as a greyscale one. We believe that using multiple colormaps will enhance the model's performance and allow it to learn from a larger data size once it has learned the features for a crack well. 
\par \noindent 
We also discuss methods we have adopted to make use of image post-processing to create samples that are harder to detect from existing samples. This involves making use of the inherent color properties of an image to help in augmentation. We can firstly use brightness which gives us the intensity of light in an image. The second is exposure, which tells us the amount of light that reaches the camera sensor. These two control how light or dark an image appears. It is interesting to note that for data generated through an experimental setup, getting images with varied values for these parameter may be difficult due to fixed sensor settings, however, we can do this very easily as a post-processing routine. The next three properties which we could also control are the saturation, contrast and the warmth. Saturation is the intensity or purity of colors in an image. Contrast refers to the difference between the darkest and the lightest parts while warmth refers to the presence of red and yellow tones in an image. Randomizing these values allow for a considerable amount of data augmentation. Further, it allows us to generate more difficult samples due to the mixing. We have shown an image being transformed so that the crack location becomes less and less obvious and one has to look at the secondary artifacts, such as the distortion created in the thermal contours by the crack in order to detect it. 
\begin{figure}[H]
\centering\includegraphics[scale = 0.3]{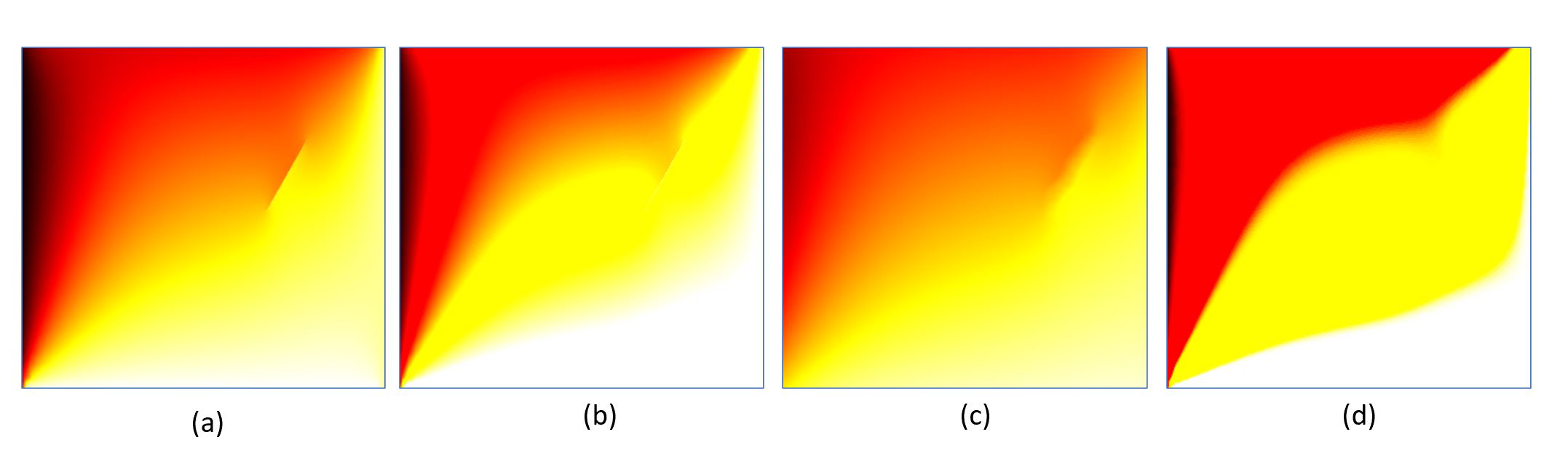}
        \caption{From left to right $\longrightarrow$ (a) Original image. (b) Image with high Brightness. (c) Image with parts near the crack blurred to decrease visibility. (d) Image with contrast, brightness and exposure adjusted such that it is harder to locate the crack}
\end{figure}
\begin{figure}[H]
    \centering
\includegraphics[width 
 = 0.4\textwidth] {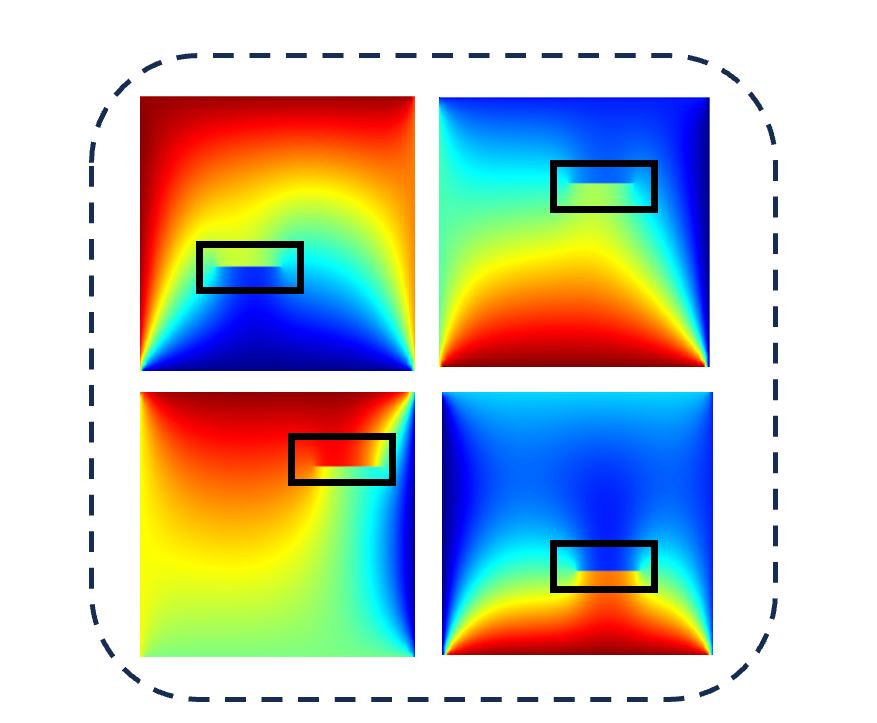}
    \caption{Easy Examples}
    \label{fig:enter-label}
\end{figure}

\begin{figure}[H]
    \centering
\includegraphics[width 
 = 0.4\textwidth] {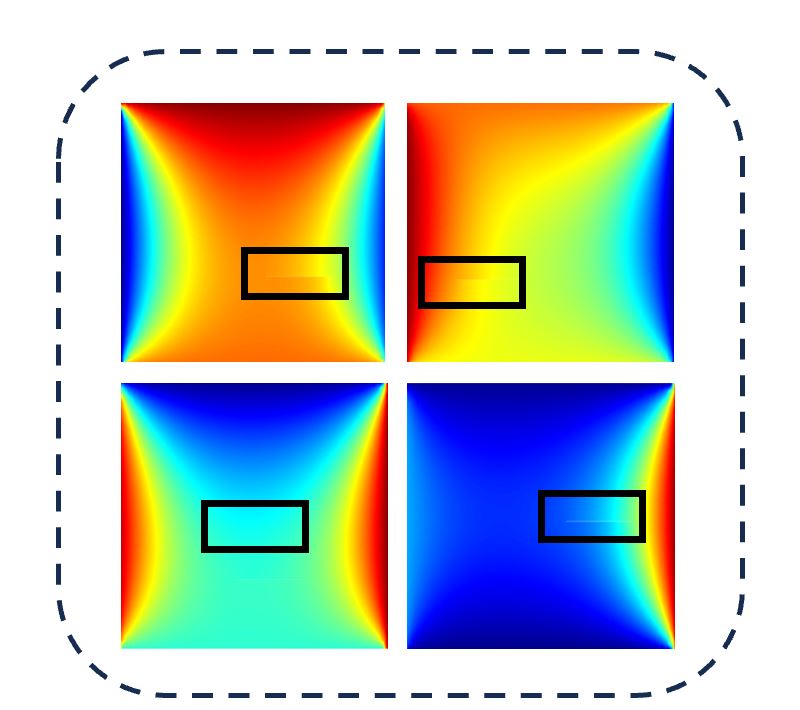}
    \caption{Hard Examples}
    \label{fig:enter-label}
\end{figure}
Easier examples are characterized by conspicuous variations in the thermal profile due to the presence of a crack. These instances are readily discernible, making it relatively straightforward for both automated systems and the human eye to identify the crack's location. On the other hand, harder examples pose a greater challenge, as the cracks are much more subtle and difficult to spot, even for the human eye. These instances demand heightened sensitivity and advanced detection techniques to accurately identify and locate the cracks within the thermal profile.
\section{Improving Data Generation Pipeline}
\begin{figure}[H]
    \centering
\includegraphics[scale = 0.3]{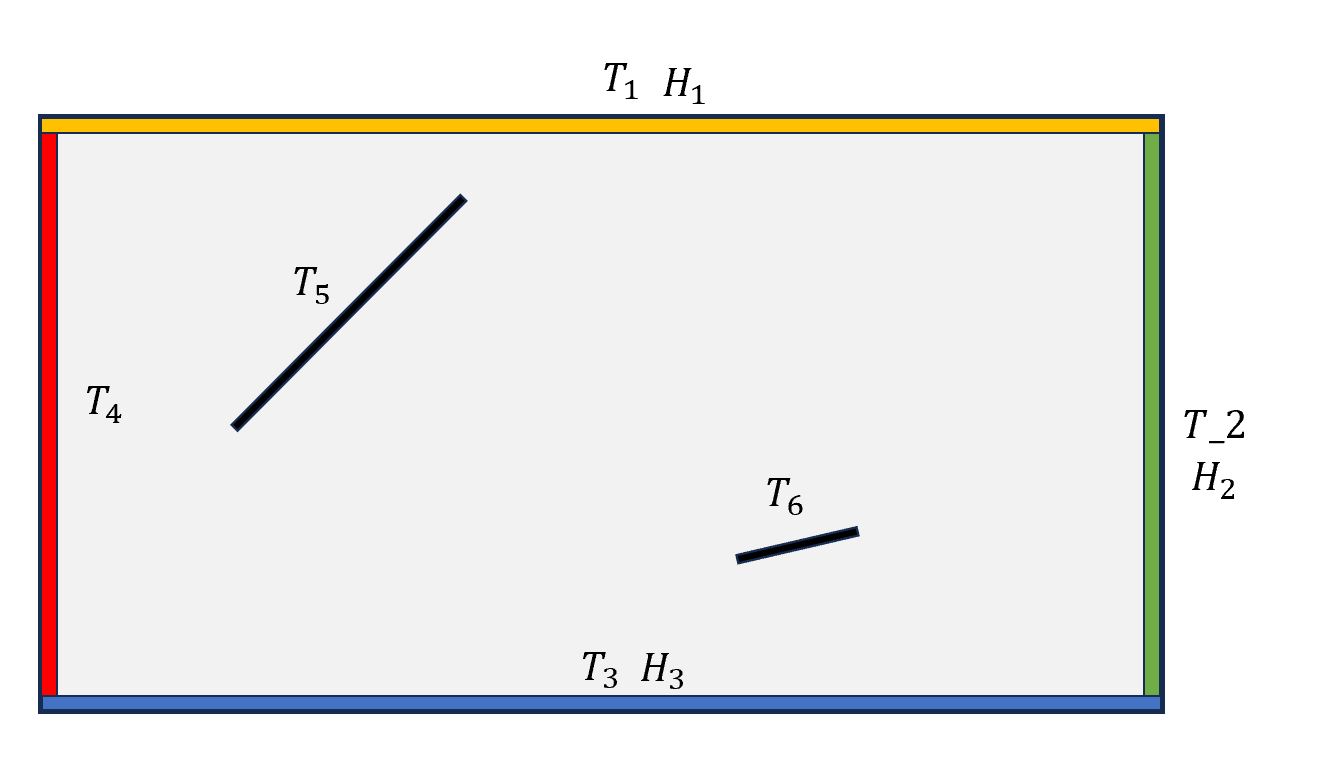}
    \caption{Improved Simulation Template}
    \label{fig:enter-label}
\end{figure}
The earlier pipeline was creating a lot of redundant images because there were only horizontal cracks. We wish to improve the data generation by applying a greater variety of boundary conditions. We also increase the diversity by randomly selecting the parameters such as the crack's center coordinates, the length of the crack, the width of the crack and the angle made by the crack with each of the plate edges.
\begin{table}[H]
\centering
\begin{tabular}{ll}
\hline
\textbf{Characteristic} & \textbf{Range} \\ \hline
Plate Length            &      4 units          \\
Plate Width             &              4 units  \\
Crack Width             &       $10^{-i}$ units $\in \{2,3,4\} $         \\
Crack Length  & $\in [0.3, 0.7]$ units \\
Crack Location & anywhere in the plate \\
Crack Inclination & $\in [0, 2\pi]$ \\
\hline
\end{tabular}
\end{table}
\begin{figure}[H]
    \centering
    \includegraphics[scale = 0.7]{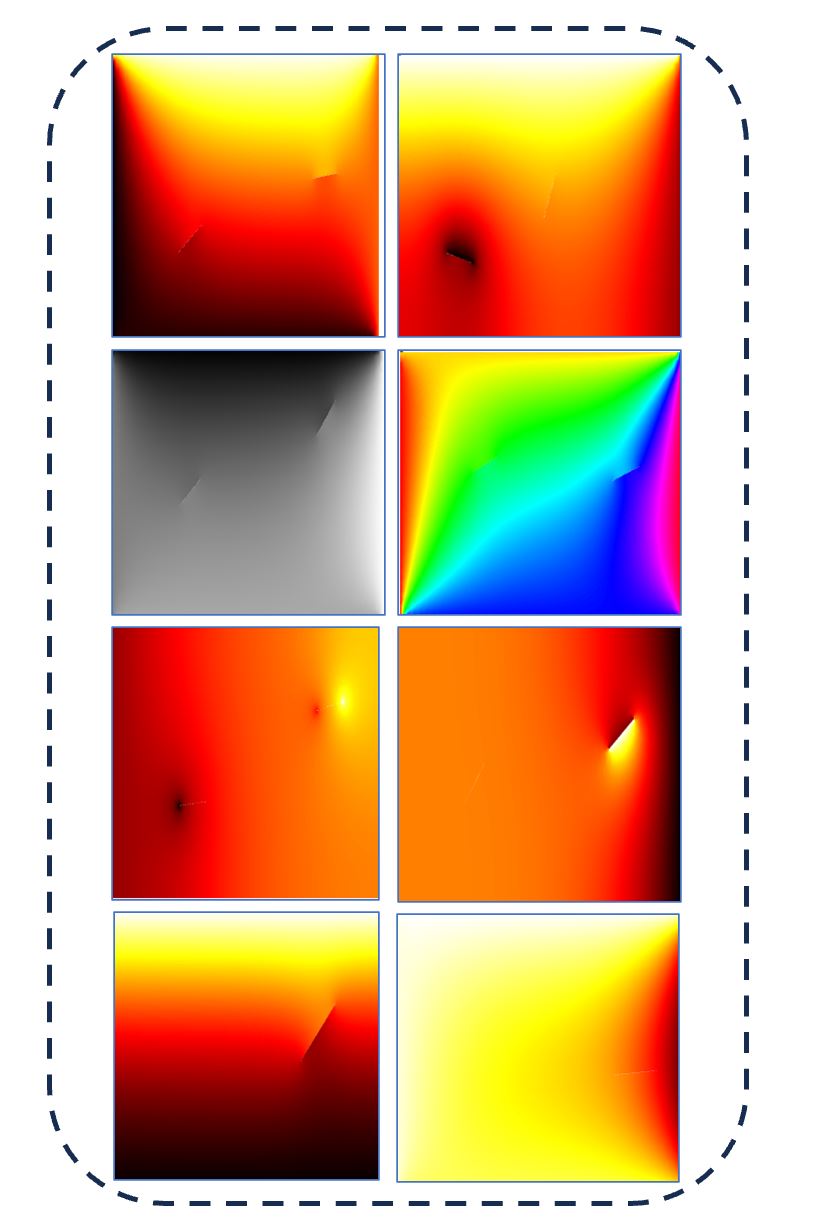}
    \caption{Examples generated by new pipeline}
    \label{fig:new_data}
\end{figure}

\begin{algorithm}
\caption{Thermal Profile Data Generation Pipeline}
\begin{algorithmic}[1]
    \STATE \textbf{Input:} Plate dimensions $L \times L$, thermal properties (thermal conductivity, specific heat)
    \STATE \textbf{Initialize:} Crack length range $(l_{\text{min}}, l_{\text{max}})$, center location range $(x_{\text{min}}, x_{\text{max}})$, $(y_{\text{min}}, y_{\text{max}})$, orientation range $[0, \pi]$, colormaps $\{C_1, C_2, \ldots, C_n\}$

    \FOR{$i = 1$ to $N$} \COMMENT{Loop to generate $N$ samples}
        \STATE Randomly sample crack length: $l \sim \text{Uniform}(l_{\text{min}}, l_{\text{max}})$
        \STATE Randomly sample crack center $(x, y) \sim \text{Uniform}(x_{\text{min}}, x_{\text{max}}) \times \text{Uniform}(y_{\text{min}}, y_{\text{max}})$
        \STATE Randomly sample crack orientation $\theta \sim \text{Uniform}(0, \pi)$
        
        \STATE Randomly select $k$ edges from the 4 plate edges
        \FOR{each selected edge}
            \STATE Randomly apply either a heat flux or constant temperature boundary condition
        \ENDFOR

        \STATE Define plate geometry with crack parameters $(l, (x, y), \theta)$

        \IF{steady-state simulation}
            \STATE Run steady-state finite element simulation
        \ELSE
            \STATE Run transient finite element simulation
        \ENDIF

        \STATE Store temperature profile image as $\text{image}_i$

        \STATE Randomly select colormap $C \in \{C_1, C_2, \ldots, C_n\}$
        \STATE Apply $C$ to $\text{image}_i$ for augmentation

        \STATE Annotate crack region (bounding box or segmentation) on $\text{image}_i$

        \STATE Save annotated image and associated data
    \ENDFOR
\end{algorithmic}
\end{algorithm}

\subsection{Deep Learning Pipeline}
\begin{figure}[H]
    \centering
\includegraphics[width 
 = 0.45\textwidth] {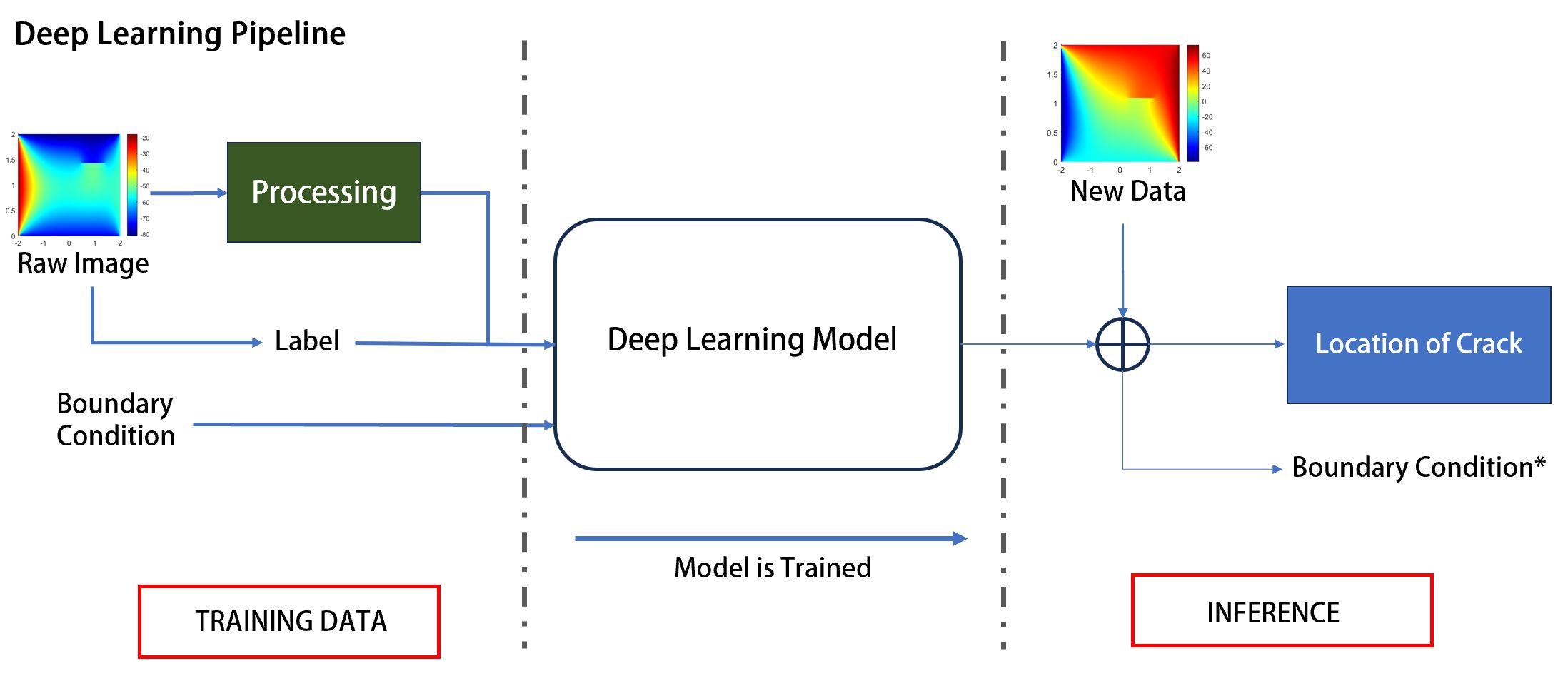}
    \caption{The Deep Learning Pipeline}
    \label{fig:enter-label}
\end{figure}

\subsection{Benchmarking}
In the process of validating the solutions, benchmarking was carried out by comparing our results with an analytical solution derived from the paper titled "Thermal Stresses In Plates with Circular Holes" by K.S. Rao et al., published in Nuclear Engineering and Design in 1971. Given that this paper focuses on a circular hole, we specifically compared our simulation methodology with the analytical solution for the same geometric configuration. Once the benchmarking process confirmed the validity of our method, simulations were extended to encompass the crack geometry, ensuring the reliability and accuracy of our approach in modeling thermal stresses.
\begin{figure}[H]
    \centering
\includegraphics[width 
 = 0.42\textwidth] {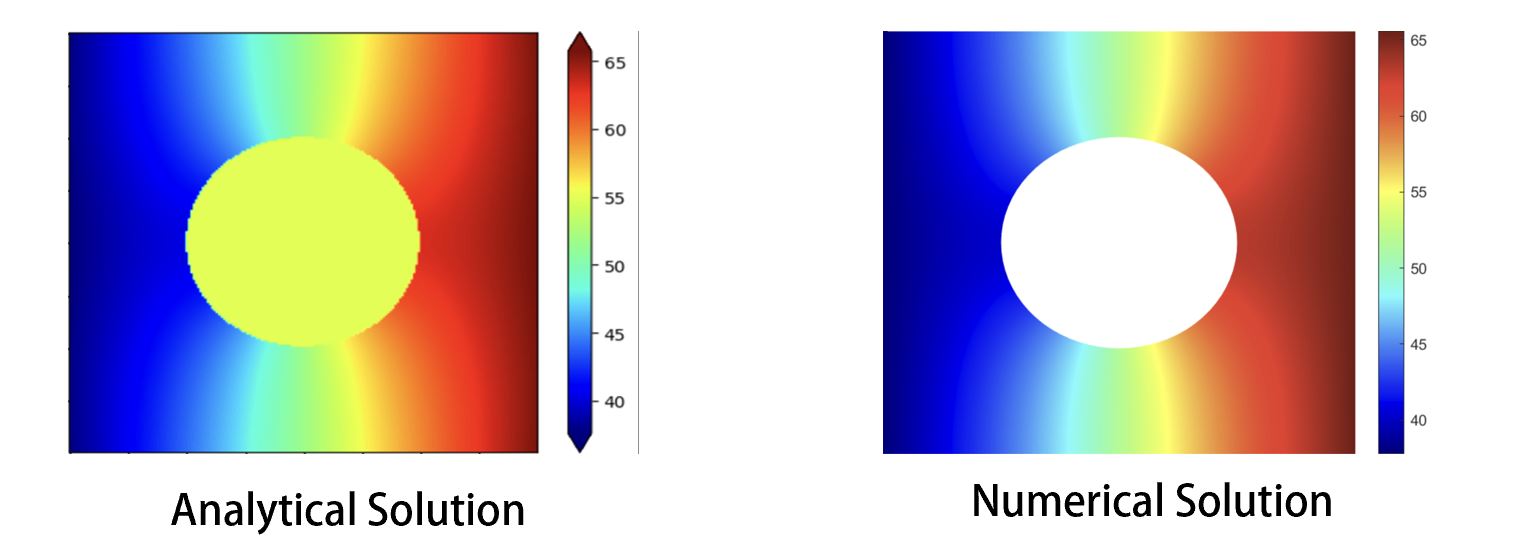}
    \caption{Benchmarking}
    \label{fig:enter-label}
\end{figure}
A thorough comparison for a circular hole geometry was conducted, revealing a high level of agreement between the analytical and numerical solutions. The visual examination indicates a close match between the two, with magnitudes exhibiting a striking similarity. It's important to note that the green circle in the analytical solution serves as a mask, signifying regions where values were not computed.
\section{Deep Learning}
\subsection{Feature Extraction}
In neural networks, the detection process involves analyzing low-level features initially, which are then integrated to identify higher-level features. The neural network focuses on detecting these features at a smaller scale and subsequently combines them using weighted operations. In our problem, just as in a majority of scenarios, the local level properties corresponding to the region where the cracks lie are much more pronounced. 
Temperature fluctuations result in rapid changes in pixel values in the proximity of a crack. This effect is more pronounced when the crack is perpendicular to the direction of maximum opposite edge temperature difference. Abrupt alterations in the direction of thermal contour lines occur as they become perpendicular to the crack. We would ideally want the neural network to able to detect all of these features and use them in the final classification process. 

\begin{figure}[H]
    \centering
\includegraphics[width 
 = 0.42\textwidth] {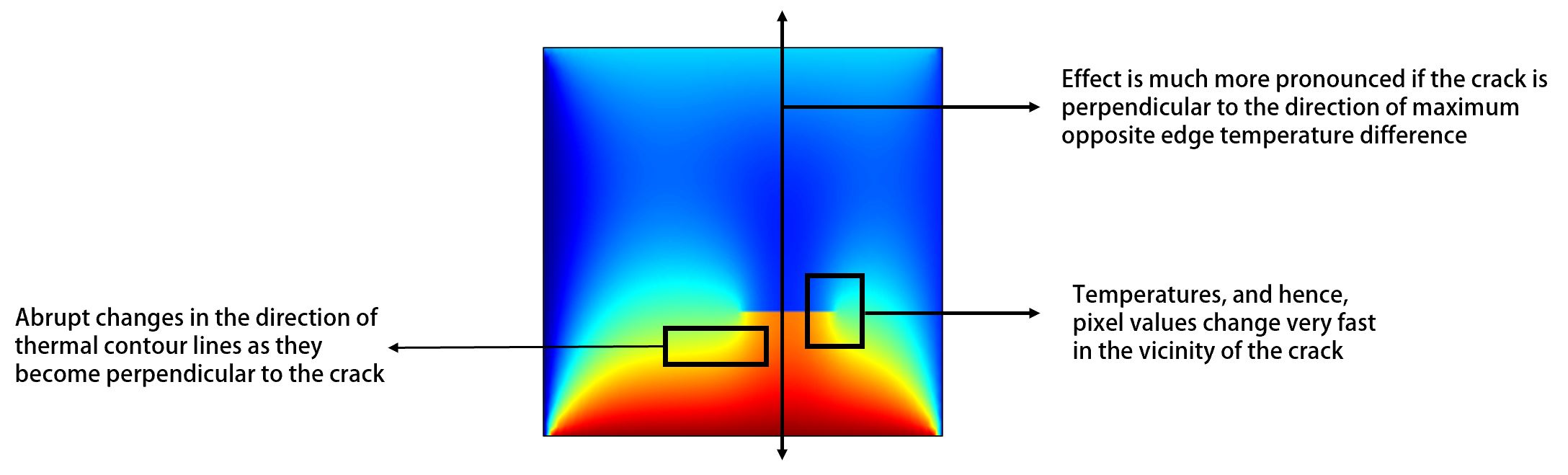}
    \caption{Local Features from Neural Networks}
    \label{fig:enter-label}
\end{figure}
\subsection{Edge Detection based on Convolution}
This is a key operation happening in the background of convolutional networks which is extremely important in our case. This is because cracks are generally similar to one-dimensional objects and represent changes high in magnitude along that direction. Therefore, using an edge detection filter can help us identify these cracks. Common edge detection filters include the Sobel, Prewitt, and Roberts operators. The below image shows the result obtained by passing an edge detection kernel over the image. 
\begin{figure}[H]
    \centering
\includegraphics[width 
 = 0.42\textwidth] {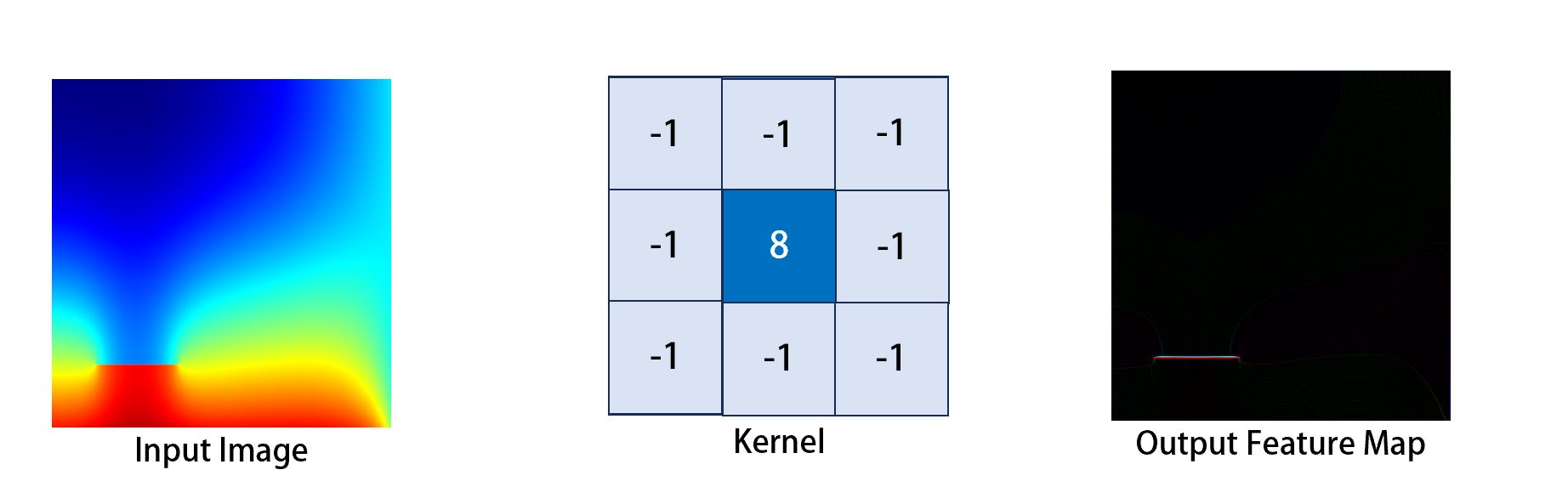}
    \caption{Edge Detection Kernel}
    \label{fig:enter-label}
\end{figure}
The edge detection kernel applied to the input image generates an output feature map that predominantly appears black, with the exception of the area corresponding to the location of the crack. 
\subsection{Architecture}
We implement RCNN models for the tasks of object detection and semantic segmentation. The YOLO series of models works very well for these tasks. We also implement the RCNN Model from the Detectron2 series from Meta AI. These models have been pre-trained on massive amounts of data. We fine-tuned these models on our synthetic datasets. 
\section{Object Detection}
\subsection{Parameters}
\begin{table}[H]
\begin{tabular}{|l|l|}
\hline
\textbf{Entity}                     & \textbf{Details}                    \\ \hline
\textbf{Size   of training dataset} & 80   images (280 x 280 px)*         \\ \hline
\textbf{Size   of testing dataset}  & 20   images (280 x 280 px)*         \\ \hline
\textbf{Optimizer}                  & Stochastic   Gradient Descent (SGD) \\ \hline
\textbf{Metrics}                    & Precision,   Recall, mAP50, mAP50-95  \\ \hline
\textbf{Number   of Epochs}         & 250                                \\ \hline
\textbf{Training   Image Resize}    & 512   px                           \\ \hline
\textbf{Batch   size}               & 8                                  \\ \hline
\textbf{Training   Time per Epoch}  & 2.22   seconds (for dataset with size 80) \\ \hline
\textbf{Inference   Time}           & 0.154   seconds (for 20 images)    \\ \hline
\end{tabular}
\end{table}
*The reason why we have used such a small number of data points is that the model was able to perform well even with a small dataset. In further experiments, we shall use augmentation to increase the dataset size. 
\subsection{Metrics Obtained}
\begin{table}[H]
\begin{tabular}{lll}
\hline
\textbf{Metric}    & \textbf{Value} & \textbf{Interpretation}                                                                      \\ \hline
\textbf{Precision} & 0.996          & Value of precision at the maxima \\ & &  of the F1-confidence curve                       \\
\textbf{Recall}    & 0.95           & Value of recall at the maxima \\ & &  of the F1-confidence curve                          \\
\textbf{mAP50}     & 0.947          & Mean Average Precision when confidence  \\ & &  threshold is set to 0.5                    \\
\textbf{mAP50-95}  & 0.507*         & mAP averaged across confidence   \\ & & thresholds from 0.5 to 0.95 with step  \\ & & size of 0.05 \\ \hline
\end{tabular}
\end{table}
We are getting very good values of precision, recall and mAP50 while mAP50-95 value is not very optimal

\subsection{Predictions}
\begin{figure}[H]
    \centering
\includegraphics[width 
 = 0.45\textwidth] {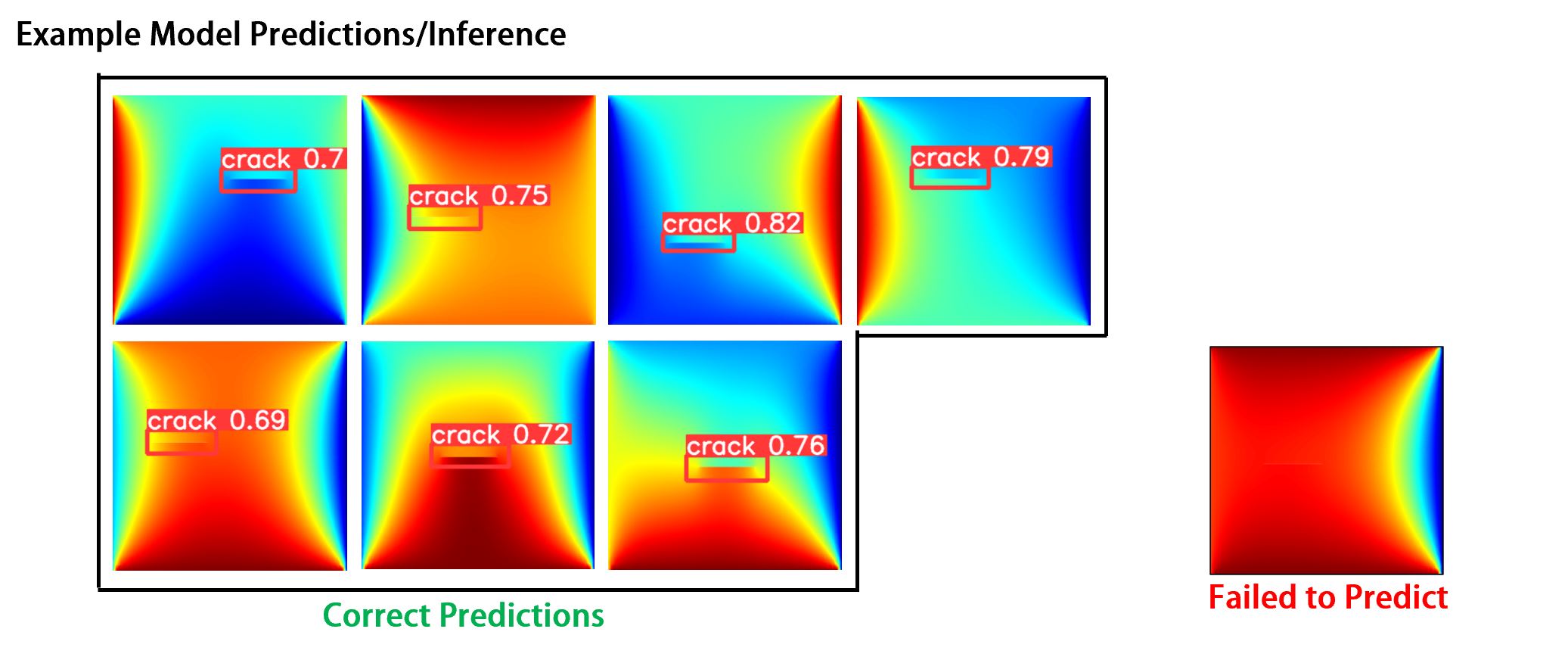}
    \caption{Predictions}
    \label{fig:enter-label}
\end{figure}
We can thus see that the model is able locate the crack correctly in most cases, but it fails during 
hard cases, when even a human eye is unable to locate the crack. 
\section{Segmentation}
Segmentation is a pivotal technique in computer vision and image processing, aiming to partition an image into meaningful regions. Various types of segmentation methods exist, each tailored to specific applications. Firstly, there's semantic segmentation, which assigns a class label to each pixel in the image, delineating objects or regions based on their semantics. Then, instance segmentation takes a step further by not only classifying each pixel but also distinguishing between individual instances of objects. This method is particularly valuable in scenarios where objects overlap or occur multiple times within an image. However, instance segmentation is computationally intensive due to the need for precise delineation, making it slower than semantic segmentation. Additionally, it requires more extensive training data and often struggles with accurately segmenting objects with intricate shapes or occlusions. The trade-offs lie in the balance between accuracy and efficiency. While instance segmentation offers granular insights into object instances, it demands more computational resources and meticulous tuning. Thus, the choice between segmentation methods hinges on the specific requirements of the task at hand, weighing the precision required against the computational constraints.
\subsection{Fine-Tuning}
Fine-tuning involves training a model that was pre-trained on some large dataset on our small dataset. This involves freezing of the initial layers of the model, while updating only the later weights. The earlier layers of any deep learning model generally capture the more generic features such as edges, curves which are not really instance-dependent and are the property of any image, not considering what the theme is actually. However, the later layers start looking at more specific features, for example, facial expression in humans or the ratios of the face and eye of an animal etc. We want the weights of these layers to adapt according to the task that we are performing, and this is done by supplying a limited amount of training data to the pre-trained model. 
\subsection{Annotation Process}
Annotation of data, especially images is generally a laborious task. In our case, for example, we need to use online tools and manually label each image by drawing a polygonal boundary around every instance of the class which we want the model to detect.
\subsection{Detectron 2}
Detectron 2, developed by Meta AI, is an advanced object detection framework known for its robustness and versatility in computer vision tasks. Building upon its predecessor, Detectron, this model boasts improvements in speed, accuracy, and flexibility. Leveraging state-of-the-art deep learning techniques, Detectron 2 provides a comprehensive toolkit for researchers and developers to tackle a wide range of challenges in object detection, instance segmentation, and related domains. Its modular architecture allows for easy customization and integration, making it a preferred choice for both academia and industry applications alike.
\subsection{Mask RCNN}
Mask R-CNN represents a paradigm shift in computer vision, seamlessly integrating object detection and instance segmentation. It extends the Faster R-CNN architecture by incorporating a parallel branch dedicated to generating pixel-level segmentation masks. Leveraging a Region Proposal Network (RPN) for object detection and a Mask branch for segmentation, Mask R-CNN achieves state-of-the-art performance in delineating object boundaries at the pixel level. This technical advancement enables precise segmentation of objects within an image, facilitating tasks such as object counting, scene understanding, and image manipulation. With its robustness and accuracy, Mask R-CNN has become a cornerstone in the field of semantic segmentation, empowering researchers and practitioners to tackle complex visual recognition challenges with unprecedented detail and precision.
\section{Results}
\subsection{Using pre-trained models directly}
We observe that if we directly use a pre-trained model without any sort of fine-tuning, the results observed are quite poor. Further, models trained on crack data, but from a different origin, such as pavement data, end up performing poorly when trying to detect the cracks from the image
\subsection{Training with Fine-tuning}
We trained a Detectron Model with the following dataset and characteristics
\begin{table}[H]
\centering
\begin{tabular}{ll}
\hline
\textbf{Dataset} & \textbf{Size} \\ \hline
Training Dataset   &     105 images     \\
Validation Dataset           &              30 images \\
Testing Dataset        &       15 images       \\
\hline
\end{tabular}
\end{table}
\section{Inference Results}
\subsection{Results on testing data from our datasets}
\begin{figure}[H]
    \centering
    \includegraphics[scale = 0.3]{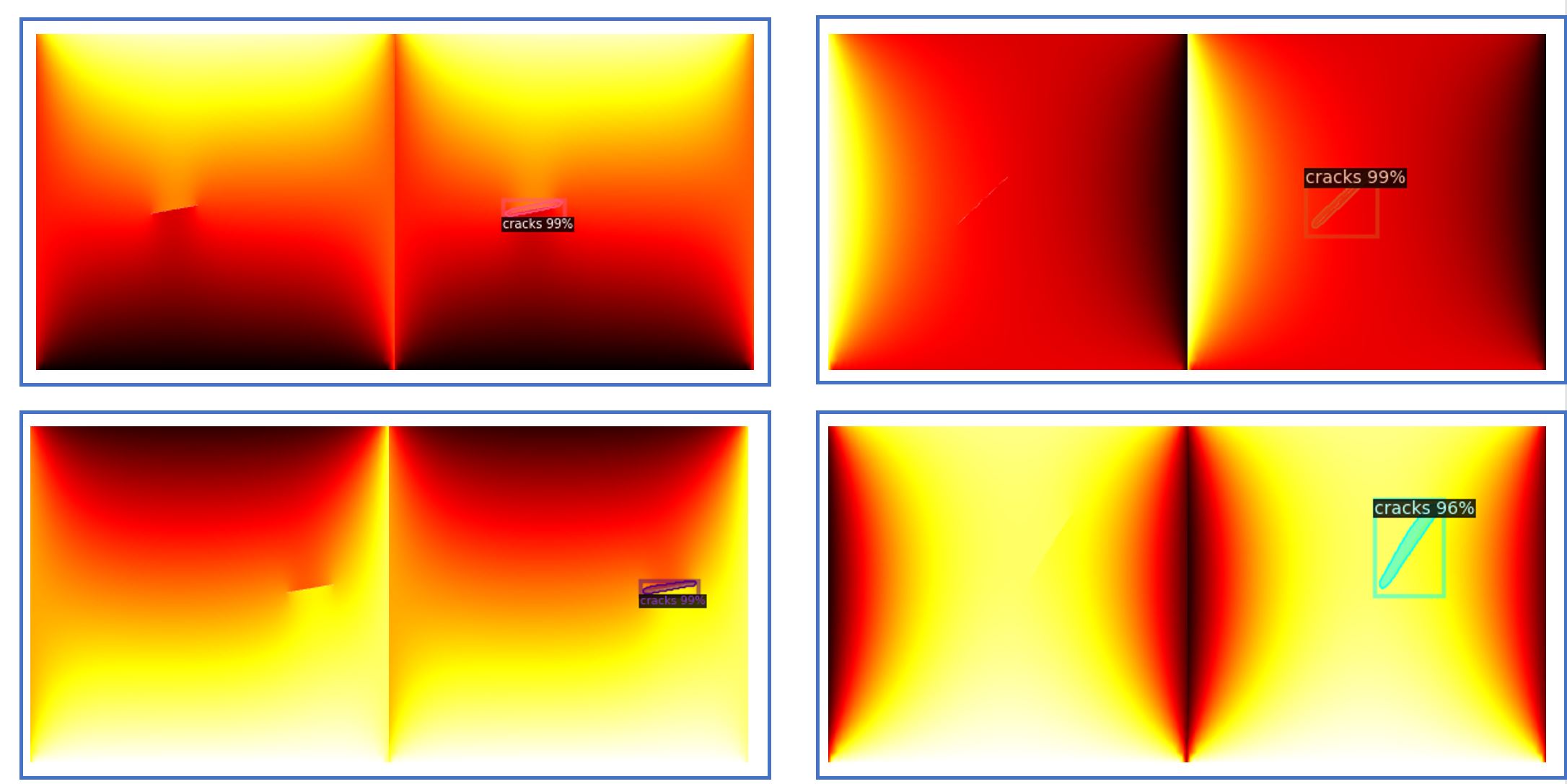}
    \caption{Inference Results}
    \label{fig:enter-label}
\end{figure}
The inference results are very promising. Here, we just grade the model on basis of whether it was able to detect the crack or not. We observe that in each of the 15/15 cases, the model was able to detect the crack correctly. This is excellent as the model is also succeeding in situations where a human has to squint a lot in order to be able to detect the image. In a couple of cases, detecting the crack was next to impossible for a human. Thus, this shows that finetuning is indeed a very good way to train a model on such a task. 
\subsection{Domain Adaptation Check: Inference Results on Images From the Wild}
In addition to testing on our own dataset, we also tried to check if the model performed well on images that were taken from sources in the wild. These images were much different in the texture and pixel resolutions as compared to the images from our training dataset. Further, these were picked up from real-world scenarios and hence other image characteristics were also very different. It has been a common observation that crack detection tasks are generally very sensitive to the differences between the training and testing data. This is something that we also observed in our experiment. Due to difficulties in data acquisition, the number of images that we could procure were of the order of $10^1$. However, there were some clear observations that we could make. 
\begin{enumerate}
    \item[(a)] The shape of the crack is extremely important. Our model was able to detect cracks in images where the shape was similar to that in the training data. The dimension is relatively less important but should not be ignored. 
    \item[(b)] The cracks should generally cover a decent fraction of the image. Occluded cracks may be difficult to detect. 
    \item[(c)] It is also very important that the texture and granularity of the images on which we test the data be very similar to that of the training data. For example, the images produced by simulations are relatively very clean and crisp. Hence, one method may be to add some kind of noise to these images. 
    \item[(d)] Since the model was trained on thermogram images, it could not scale onto crack detection scenarios where the images are simply grayscale. 
\end{enumerate}
Although there is scope for improvement, the results are promising. Using a more diverse and varied dataset will surely lead to much better results since we do observe domain adaptation in some cases. Similarly, fine-tuning a network on our dataset and then again performing some form of transfer learning also holds promise. 
\begin{table*}[t]
\centering
\begin{tabular}{|l|c|c|c|}
\hline
\textbf{Does model scale well to Test Data?}       & Highly Similar to Training Data & Somewhat Similar to Training Data & Vastly Different from Training Data \\ \hline
\textbf{Shape of crack}              & \ding{51}         & \ding{55}         & \ding{55}         \\ \hline
\textbf{Size of crack}              & \ding{51}         & \ding{51}         & \ding{55}         \\ \hline
\textbf{Texture of Image}              & \ding{51}         & \ding{55}         & \ding{55}         \\ \hline
\textbf{Thermal Contours}              & \ding{51}         & \ding{51}         & \ding{55}         \\ \hline
\textbf{Foreground-Background Differences}              & \ding{51}         & \ding{55}         & \ding{55}         \\ \hline
\end{tabular}
\caption{Analysis of whether the model is able to adapt to new domains. \\ \ding{51} represents good performance when that condition is satisfied while \ding{55} represents bad performance.}

\end{table*}
\section{Possible Improvements for Vision Tasks}
The possible improvements for the data generation process have been mentioned in the future scope in detail. Here, we focus on the improvements that are possible from the deep learning side. We would like to mention that the primary goal of our project was to show how a data generation pipeline along with deep learning can work well in practice. We have not therefore tried to match any SOTA performances. If one were to do so, a possible improvement would have been to experiment on a larger number of model architectures and backbones. These models could also be selected on the exact use case, taking into account other factors such as inference time and training complexity and parameter size as well. The second possible improvement is the tuning of model hyperparameters. This is not of that much significance due to the use of a pre-trained backbone. However, optimization of the model hyperparameters may yield in improvements in the results. 
\section{Future Scope}
Our approach shows that training an end-to-end model on data generated through such a synthetic process is indeed a viable option for a large number of practical applications. We have quite a few future avenues that can be discussed and explored as an extension to this work. Specifically, focusing on the problem of crack detection in steel plates, one can first define the crack in various ways. The current methodology adopted through our approach is to define a crack as a region from which material is removed; however, one could also have cracks as a region where there are changes in the material's mechanical properties. Indeed, the detection of cracks in such a scenario might be an even more difficult problem. Our approach considers mostly cracks which are on or close to the surface of a material. However, there is also the problem of sub-surface crack detection. In such scenarios, a possible approach is to obtain images of the surface by slicing at different parts through the depth of the steel object and then reconstructing these slices to create a three-dimensional image object. This can also be treated as a sequential problem. Further, one must also study the sensitivity of the surface temperature to the temperature profiles at a sub-surface level in order to find out if we can detect sub-surface cracks via an analysis of the surface-level temperature plots. \par 
The next possible approach is use more complex artificial intelligence models to use the synthesized images as a base and then build up on top of them. To increase the amount of complexity, one can also use methods such as Style Transfer from experimental images and use them on the data which we have created synthetically. The performance of methods on such an augmented dataset is worth checking. Further, it is also possible to use conditional generative models that use the synthetically generated data as a prior and create images pertaining to more complicated simulations. This will allow us to create more data in a tunable manner. Further, we can also explore different kinds of experimental settings for the finite element simulations. Different kinds of softwares will indeed lead to an increase in the data formed as well. We can currently very easily adapt to different kinds of two-dimensional geometries with the pipeline we have, however, it is also interesting to see how we can make use of three-dimensional geometries for the same problem since that will lead to a lot of interesting scenarios which cannot be explored in the two-dimensional setting. Further, we can make use of different angles from which images are taken similar to how it is done in a real-life scenario. \par In addition to these things, we also need to find ways of performing effective domain adaptation. The diversity in the data created has to be high to ensure proper translation to new types of images. This can be achieved via the creation of datasets having a very high level of variety. How our created images directly map to the images obtained through experimental images is also an avenue to check in the future. This can be achieved via the use of contrastive learning methods which evaluate some sort of a distance metric between the created images and the real images. This can also be achieved by using some sort of a Generator-Discriminator framework such as a GAN. Further, we could also make use of a Variational Autoencoder for which the latent embeddings are passed via the synthetic datasets.


\begin{thebibliography}{00}
\bibitem{b1} Jaeger, B.E., Schmid, S., Grosse, C.U. et al. Infrared Thermal Imaging-Based Turbine Blade Crack Classification Using Deep Learning. J Nondestruct Eval 41, 74 (2022). https://doi.org/10.1007/s10921-022-00907-9
\bibitem{b2} Yang, J., Wang, W., Lin, G., Li, Q., Sun, Y., \& Sun, Y. (2019). Infrared Thermal Imaging-Based Crack Detection Using Deep Learning. IEEE Access, 7, 182060-182077.
\bibitem{b3} Arun Mohan, Sumathi Poobal,
Crack detection using image processing: A critical review and analysis,
Alexandria Engineering Journal,
Volume 57, Issue 2,
2018,
Pages 787-798,
ISSN 1110-0168,
https://doi.org/10.1016/j.aej.2017.01.020.
\bibitem{b4} Tian, L., Wang, Z., Liu, W. et al. A New GAN-Based Approach to Data Augmentation and Image Segmentation for Crack Detection in Thermal Imaging Tests. Cogn Comput 13, 1263–1273 (2021). https://doi.org/10.1007/s12559-021-09922-w
\bibitem{b5} Alexander, Q.G., Hoskere, V., Narazaki, Y. et al. Fusion of thermal and RGB images for automated deep learning based crack detection in civil infrastructure. AI Civ. Eng. 1, 3 (2022). https://doi.org/10.1007/s43503-022-00002-y
\bibitem{b6} Chandra S, AlMansoor K, Chen C, Shi Y, Seo H. Deep Learning Based Infrared Thermal Image Analysis of Complex Pavement Defect Conditions Considering Seasonal Effect. Sensors. 2022; 22(23):9365. https://doi.org/10.3390/s22239365
\bibitem{b7} Péter Kovács, Bernhard Lehner, Gregor Thummerer, Günther Mayr, Peter Burgholzer, Mario Huemer; Deep learning approaches for thermographic imaging. J. Appl. Phys. 21 October 2020; 128 (15): 155103. https://doi.org/10.1063/5.0020404
\bibitem{b8} Fang Q, Ibarra-Castanedo C, Maldague X. Automatic Defects Segmentation and Identification by Deep Learning Algorithm with Pulsed Thermography: Synthetic and Experimental Data. Big Data and Cognitive Computing. 2021; 5(1):9. https://doi.org/10.3390/bdcc5010009
\bibitem{b9} Matlab Official Documentation: https://in.mathworks.com/help/matlab/
\bibitem{b10} YOLOv5 Official Implementation: https://github.com/ultralytics/yolov5
\bibitem{b11}
Wood, E., Baltrušaitis, T., Hewitt, C., Dziadzio, S., Cashman, T. J., \& Shotton, J. (2021). Fake it till you make it: face analysis in the wild using synthetic data alone. Proceedings of the IEEE/CVF International Conference on Computer Vision, 3681–3691.
\bibitem{b12} He, R., Sun, S., Yu, X., Xue, C., Zhang, W., Torr, P., Bai, S., \& Qi, X. (2022). Is synthetic data from generative models ready for image recognition? ArXiv Preprint ArXiv:2210.07574.
\bibitem{b13} Nikolenko, S. I. (2021). Synthetic data for deep learning (Vol. 174). Springer.
\bibitem{b14} Wang, Q., Gao, J., Lin, W., \& Yuan, Y. (2021). Pixel-wise crowd understanding via synthetic data. International Journal of Computer Vision, 129(1), 225–245.
\bibitem{b15} Hahner, M., Sakaridis, C., Dai, D., \& Van Gool, L. (2021). Fog Simulation on Real LiDAR Point Clouds for 3D Object Detection in Adverse Weather. Proceedings of the IEEE/CVF International Conference on Computer Vision (ICCV), 15283–15292.
\bibitem{b16} Lu, Y., Shen, M., Wang, H., Wang, X., van Rechem, C., Fu, T., \& Wei, W. (2023). Machine learning for synthetic data generation: a review. ArXiv Preprint ArXiv:2302.04062.
\bibitem{b17} Roberts, M., Ramapuram, J., Ranjan, A., Kumar, A., Bautista, M. A., Paczan, N., Webb, R., \
\& Susskind, J. M. (2021). Hypersim: A Photorealistic Synthetic Dataset for Holistic Indoor Scene Understanding. Proceedings of the IEEE/CVF International Conference on Computer Vision (ICCV), 10912–10922.
\bibitem{b18} Wu, X., Hong, D., \& Chanussot, J. (2022). UIU-Net: U-Net in U-Net for infrared small object detection. IEEE Transactions on Image Processing, 32, 364–376.
\bibitem{b19} Tokmakov, P., Li, J., Burgard, W., \& Gaidon, A. (2021). Learning To Track With Object Permanence. Proceedings of the IEEE/CVF International Conference on Computer Vision (ICCV), 10860–10869.
\bibitem{b20} Jain, S., Seth, G., Paruthi, A., Soni, U., \& Kumar, G. (2022). Synthetic data augmentation for surface defect detection and classification using deep learning. Journal of Intelligent Manufacturing, 1–14.

\end{thebibliography}
\end{document}